\documentclass{article}
\usepackage{ijcai26}

\usepackage{times}
\usepackage{soul}
\usepackage{url}
\usepackage[hidelinks]{hyperref}
\usepackage[utf8]{inputenc}
\usepackage[small]{caption}
\usepackage{graphicx}
\usepackage{amsmath}
\usepackage{amsthm}
\usepackage{booktabs}
\usepackage{algorithm}
\usepackage{algorithmic}
\usepackage[switch]{lineno}

\usepackage[table]{xcolor}
\usepackage{xcolor}
\usepackage{threeparttable}
\usepackage{tikz}
\usepackage{amssymb}

\title{Towards Automated Kernel Generation in the Era of LLMs}

\author{
Yang Yu$^1$
\and
Peiyu Zang$^{1,2}$
\and
Chi Hsu Tsai$^{1,3}$\and
Haiming Wu$^{1,4}$\and
Yixin Shen$^{1,5}$\and \\
Jialing Zhang$^{1,6}$\and
Haoyu Wang$^{1,7}$\and 
Zhiyou Xiao$^{1,3}$\and
Jingze Shi$^{8}$\and
Yuyu Luo$^{8}$\and \\
Wentao Zhang$^{3}$\and
Chunlei Men$^{1}$\and
Guang Liu$^{1}$ \And
Yonghua Lin$^1$\\
\affiliations
$^1$Beijing Academy of Artificial Intelligence \\
$^2$Beijing Normal University \\
$^3$Peking University \\
$^4$Beijing Institute of Technology \\
$^5$Cornell University \\
$^6$Beijing Jiaotong University \\
$^7$Renmin University of China\\
$^8$Hong Kong University of Science and Technology (Guangzhou)\\
}

\begin{document}

\maketitle

\begin{abstract}
    The performance of modern AI systems is fundamentally constrained by the quality of their underlying GPU kernels, which translate high-level algorithmic semantics into low-level hardware operations. Achieving near-optimal kernels requires expert-level understanding of hardware architectures and programming models, making kernel engineering a critical but notoriously time-consuming and non-scalable process. Recent advances in large language models and LLM-based agents have opened new possibilities for automating kernel generation and optimization. LLMs are well-suited to compress expert-level kernel knowledge that is difficult to formalize, while agentic systems further enable scalable optimization by casting kernel development as an iterative, feedback-driven loop. Rapid progress has been made in this area. However, the field remains fragmented and lacks a systematic perspective for LLM-driven kernel generation. This survey addresses this gap by providing a structured overview of existing approaches, spanning LLM-based approaches and agentic optimization workflows, and systematically organizing the datasets and benchmarks that underpin learning and evaluation in this domain. Moreover, key open challenges and future research directions are further outlined, aiming to establish a comprehensive reference for the next generation of automated kernel optimization. To keep track of this field, we maintain an open-source GitHub repository at \url{https://github.com/flagos-ai/awesome-LLM-driven-kernel-generation}.
\end{abstract}

\section{Introduction} \label{sec:intro}

Rapid scaling of large language models (LLMs) has made efficient hardware utilization a central challenge in modern AI systems~\cite{kaplan2020scaling}. Consequently, specialized accelerators such as GPUs and NPUs have become the foundation of large-scale training and inference~\cite{choquette2021nvidia,huawei2019ascend}. Their performance is largely governed by kernels implementing fundamental operations, such as matrix multiplication and attention, which dominate execution time in LLM workloads. As a result, overall system throughput, efficiency, and cost are often determined more by kernel quality than by hardware peak performance.

Despite their foundational role, the development of efficient kernels remains a formidable engineering challenge. Achieving near-peak hardware utilization requires deep expertise in both algorithmic design and hardware-specific intricacies. Furthermore, kernel optimization is inherently non-scalable: implementations are often tightly coupled to particular hardware architectures and workload characteristics, which hinders their reuse and generalization across different GPU generations or hardware vendors~\cite{wu2023pytorch}. Although compiler-based autotuning approaches partially alleviate these scalability challenges through automated schedule generation and optimization, they remain fundamentally constrained by manually designed search spaces, scheduling primitives, and optimization priors. 

\begin{figure*}
    \centering
    \includegraphics[width=0.8\linewidth]{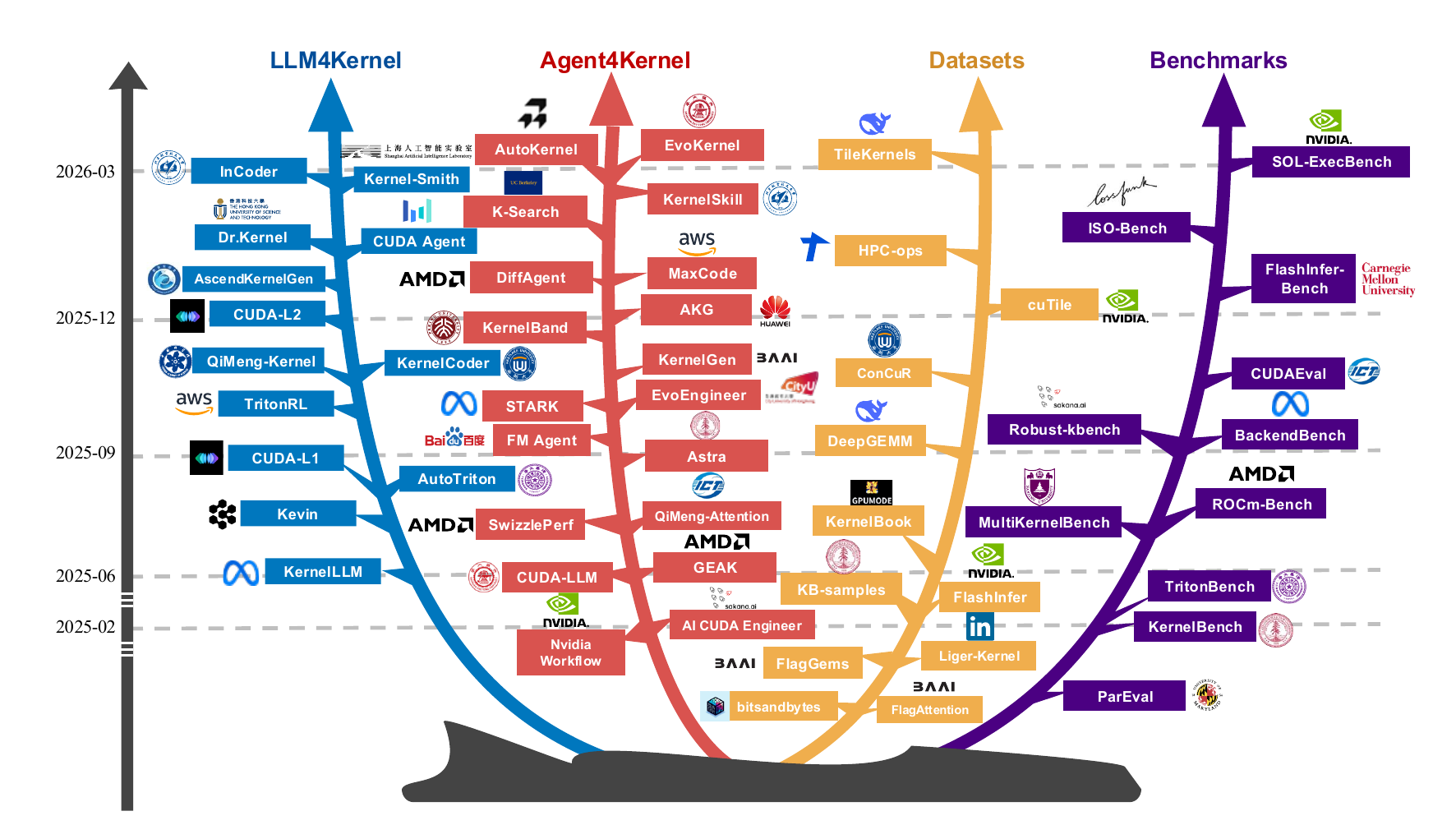}
    \caption{Illustration of the growth trend in the field of LLM-driven kernel generation. We organize these research works chronologically and categorically based on their publication dates and the domains they belong to.}
    \label{fig:1}
\end{figure*}

In response to these challenges, LLMs and agentic systems have emerged as a promising paradigm for kernel generation and optimization. Trained on large-scale code repositories and technical documentation, LLMs encode substantial expertise in hardware-aware programming, enabling them to bridge the gap between high-level algorithmic specifications and low-level implementations. Beyond one-shot code generation, agentic frameworks leverage iterative execution feedback to refine candidate kernels, facilitating scalable and  more open-ended exploration of complex optimization spaces across workloads and hardware platforms. Consequently, LLMs and agents are increasingly becoming a foundation for automated kernel development.

Despite rapid progress, research on LLM-driven kernel generation remains fragmented and lacks a systematic synthesis. This survey addresses this gap by presenting a unified overview of the field, clarifying foundational concepts, and highlighting emerging methodologies and trends. A key contribution is our consolidated resource infrastructure, which features a structured organization of training-ready kernel datasets and a literature collection tailored for retrieval-augmented generation (RAG), designed to facilitate data-driven research in this specialized kernel-generation domain. Moving beyond a synthesis of existing methodologies, we also highlight critical open challenges and propose promising research directions, aiming to establish a foundational reference for the next generation of innovation in LLM-driven kernel generation.

\section{Background}
\label{sec:background}

\paragraph{LLMs and LLM-based Autonomous Agents.} The foundation of modern LLMs is the Transformer architecture~\cite{vaswani2017attention}, which functions as the probabilistic predictor trained via the next token prediction objective. Given a sequence of tokens $x = (x_1, \dots, x_T)$, the model maximizes the joint probability:
$$
P(x) = \prod_{t=1}^{T} P(x_t \mid x_{1}, \dots, x_{t-1}; \theta).
$$
This objective enables the model to internalize world knowledge and reasoning patterns implicitly during pretraining.

While LLMs serve as the cognitive engine for reasoning and decision-making, autonomous agents extend this capability by integrating additional system components such as planning, memory, and tool-use mechanisms, and interact with the environment through trial and error \cite{wang2024survey}. In this framework, the LLM functions as the ``brain'', orchestrating actions through reasoning strategies. And agents utilize tools to perform actions beyond the model's internal knowledge and receive environmental feedback.

\paragraph{Kernel Programming and Code Generation.}

Kernel generation and optimization have traditionally followed two paradigms. The first relies on expert-written kernels and domain-specific abstractions, such as CUDA, CUTLASS, and TileLang, which enable fine-grained hardware optimization but require substantial architecture-specific expertise. The second centers on compiler-driven frameworks, such as Halide and TVM, which optimize kernels through scheduling and autotuning. While improving programmability, these approaches remain constrained by predefined search spaces, scheduling primitives, and handcrafted optimization rules. In contrast, LLM-based agents leverage knowledge distilled from large-scale code corpora and execution feedback to enable scalable and open-ended kernel optimization.

In parallel, LLMs have advanced code generation from code completion to complex software engineering tasks. However, kernel generation differs fundamentally from general-purpose code synthesis: beyond functional correctness, it must satisfy stringent performance requirements and adapt to hardware-specific execution characteristics. Consequently, it is more closely aligned with performance-oriented program synthesis and compiler optimization, necessitating specialized techniques beyond generic LLM-based code generation. This trend is also emerging in broader parallel programming systems, where LLMs have been applied to performance-oriented optimization beyond kernel generation~\cite{wei2025improving}.

\section{LLM for Kernels Generation}
\label{sec:posttraining}
Building on advances in LLM-driven code generation, recent work has increasingly applied LLMs to the generation of high-performance GPU kernels. To highlight the methodological patterns that have emerged across this landscape, the following sections review two dominant families of post-training techniques used to specialize LLMs for kernel generation: \emph{supervised fine-tuning} and \emph{reinforcement learning}.

\subsection{Supervised Fine-Tuning}

Supervised fine-tuning (SFT) has become a central methodology for enabling LLMs to synthesize high-quality kernels, relying on paired datasets that capture both high-level computational intent and low-level kernel implementation patterns. KernelLLM~\cite{kernelllm2025} adopts this strategy by collecting samples and using the Triton compiler to produce aligned PyTorch–Triton examples, and applies instruction tuning with structured prompts that explicitly encode the mapping between computation and kernel structure.  And ConCuR~\cite{DBLP:journals/corr/abs-2510-07356} further generates and curates high-quality kernel datasets with reasoning traces, motivated by the observation that the structure and clarity of model reasoning can strongly affect kernel correctness and performance. Fine-tuning on such data leads to KernelCoder, a model capable of generating CUDA kernels that achieve 17\% on the fast$_1$ metric of KernelBench~\cite{ouyang2025kernelbench} level 1. Moreover, to address the stringent demands of industrial software development, InCoder-32B~\cite{yang2026incoder}  introduces a three-stage data curation pipeline consisting of pre-training, mid-training, and post-training, achieving a fast$_1$ score of 22.2\% on KernelBench Level 1.

\subsection{Reinforcement Learning}

Reinforcement learning further enhances kernel generation via iterative feedback. Kevin~\cite{baronio2025kevin} models kernel generation as a multi-turn optimization using cross-turn reward attribution for long-horizon credit assignment. CUDA-L1 introduces contrastive RL with an LLM-as-a-judge for dense feedback, and is refined by CUDA-L2~\cite{su2025cudal2surpassingcublasperformance}, which reports performance improvements over cuBLAS on its evaluated workloads. SparseRL~\cite{wang2026mastering} leverages RL to generate high-performance CUDA code for sparse matrix operations. CUDA Agent~\cite{dai2026cuda} introduces a large-scale agentic reinforcement learning system, including a skill-augmented CUDA development environment with automated verification and profiling to provide reliable reward signals, achieving state-of-the-art results on KernelBench, delivering a 99\% faster rate over PyTorch Eager on KernelBench Level-1.

In addition to CUDA kernel generation, recent studies have investigated RL-based approaches for Triton kernel generation. AutoTriton~\cite{li2025autotriton} generates Triton kernel with the model trained by RL and addresses reward sparsity by combining structural assessments of generated kernels with execution-based runtime rewards, while TritonRL~\cite{woo2025tritonrl} extends this line of work through hierarchical reward decomposition and explicit verification of code outputs and intermediate reasoning traces. QiMeng-Kernel~\cite{zhu2025qimengkernel} further structures optimization by applying RL hierarchically to macro-thinking strategies rather than low-level implementation. Dr. Kernel~\cite{liu2026dr} introduces a robust distributed GPU environment for Triton kernel generations, combined with effective multi-turn RL methods to address the biased policy gradient issue and alleviates lazy optimization problem. Kernel-Smith~\cite{du2026kernel} presents a stable evolution-oriented  post-training recipe, and the Triton kernels generated by this framework achieve a 70\% fast$_1$ score on KernelBench level 1.
Finally, AscendKernelGen~\cite{cao2026ascendkernelgen} expands the alignment learning paradigm to Ascend NPUs and generates AscendC kernels, combining CoT-based SFT with DPO (Direct Preference Optimization).

\section{LLM Agent for Kernels Generation}
\label{sec:agents}

Relying on foundational LLMs alone typically reduces kernel development to a static one-pass inference process. In contrast, LLM-based agents introduce autonomy and feedback into the optimization loop by enabling planning, tool use, and evaluation of intermediate results. This closed-loop, self-improving paradigm allows agent-based approaches to scale kernel optimization across diverse workloads and hardware platforms, while sustaining long-horizon, fatigue-free exploration. Concretely, we categorized recent agent-driven advancements into four structural dimensions: \emph{learning mechanisms}, \emph{external memory management}, \emph{hardware profiling integration}, and \emph{multi-agent orchestration}.

\subsection{Learning Mechanisms}

This category studies how LLM agents interact with execution environments and learn effective kernel generation strategies through trial-and-error. Initial approaches view kernel generation as iterative refinement. Caesar in KernelBench utilizes simple feedback loops to refine kernels, while Inference-Time Scaling~\cite{chen2025deepseekr1} demonstrates that scaling test-time compute and reflection significantly boost kernel quality. To manage complexity, PEAK~\cite{tariq2025peak} employs a modular iterative stepwise refinement strategy. AutoKernel~\cite{jaber2026autokernel} profiles the full model to identify computational bottlenecks and employs an iterative optimization process to refine the corresponding kernel implementations.
MaxCode~\cite{ou2026maxcode} further unifies existing iterative search methods under a max-reward reinforcement learning framework, combined with a natural language critique model converting raw execution feedback into diagnostic insights. K-Search~\cite{cao2026k} leverages LLMs as world models, utilizing their prior domain knowledge to guide the search process. It decouples high-level algorithmic planning from low-level program instantiation, and demonstrates strong performance on diverse and complex kernels.

Furthermore, a large body of agent-based methods employs iterative refinement for GPU kernel generation targeting diverse model architectures, programming languages, and hardware backends. For example, DiffAgent~\cite{zhu2026diffbench} adopts iterative refinement to accelerate diffusion models, TritonX~\cite{hammond2025agentic} uses iterative refinement within a state machine to cover kernels of complete PyTorch ATen backends, and KernelGen~\cite{kernelgen} leverages test-time scaling and reflection techniques to enable kernel generation for multi-chip backends. Moreover, many existing CLI-based agents such as Claude Code or OpenCode also support iterative refinement, and AKO~\cite{ako2026} provides harnesses for such coding agents to enable agentic kernel optimization and achieve state-of-the-art performance on challenging benchmarks.

To escape local optima, recent frameworks adopt population-based evolution. Lange et al.~\cite{lange2025towards} optimize translation CUDA via mutation and crossover. FM Agent \cite{li2025fm} includes an evolutionary stage with the principles of diversity preservation, adaptive evolution, and multi-population dynamics. Advanced population dynamics are also introduced in EvoEngineer~\cite{Guo2025EvoEngineer}, which decouples traversal techniques from population management. GPU Kernel Scientist ~\cite{Andrews2025GPUKernelScientist} employs a multi-stage evolutionary workflow to address the challenge of optimizing HIP kernels for the AMD accelerators. And cuPilot~\cite{chen2025cupilot} guides evolution through high-level semantic strategies.

\subsection{External Memory Management}

Complex kernel optimization often requires domain-specific knowledge, such as CUDA APIs and hardware instruction sets that may be hallucinated or forgotten by the LLM. Agents in this category augment generation with relevant skills, domain knowledge, or prior interaction experiences, which serve as an external memory for LLMs. KernelEvolve~\cite{liao2025kernelevolvescalingagentickernel} advances the external knowledge management paradigm by integrating a sophisticated hardware-specific knowledge base specifically tailored for heterogeneous AI accelerators.
Beyond retrieving unstructured textual context, recent work has explored utilizing structured representations as external memory to guide model inference. Work such as ReGraphT \cite{gong2025large} proposes a novel framework that treats a reasoning graph as a domain-specific external memory for CUDA code optimization. In this approach, the logical transitions between optimization states of large language models are externalized into a static, navigable graph structure for the small language model to retrieve. KernelBlaster~\cite{dong2026kernelblaster} enables agents to learn from experience by accumulating knowledge into a retrievable persistent CUDA knowledge base, EvoKernel~\cite{zheng2026towards} further introduces a value-driven memory and retrieved prior experiences or trajectories with different priorities. KernelSkill~\cite{sun2026kernelskill} provides a  dual-level memory architecture, where reusable expert skills are curated in the long-term memory.

\subsection{Hardware Profiling Integration}

The third dimension addresses the hardware-agnostic nature of standard LLMs by configuring the agent's persona profile with hardware specifications, and iteratively reasoning over performance profiling feedback. 

QiMeng-TensorOp~\cite{zhang2025qimeng} triggers LLMs to analyze and distill low-level hardware documentation according to user input into the generation prompt, while QiMeng-GEMM~\cite{zhou2025qimenggemm} generates General Matrix Multiplication (GEMM) with the meta-prompt, which offers universal templates for various general optimization techniques and platform-specific optimization details. QiMeng-Attention~\cite{zhou2025qimeng} considers target GPU architecture and instruction set to convert the high-level thinking language into low-level CUDA code, and  implements the high-performance FlashAttention on different GPUs. SwizzlePerf \cite{tschand2025swizzleperf} explicitly tackles the swizzling problem, which explicitly injects precise architectural specifications into the prompt context and restricts the search space specifically to swizzling patterns that focus solely on maximizing the L2 cache hit rate.

Complementing this, agents leverage dynamic feedback. CUDA-LLM \cite{chen2025cuda} incorporates detailed target GPU specifications (e.g., warp size, cache size) into the agent's prompt. Simultaneously, compilation logs and runtime performance metrics are also aggregated to guide the optimization process. TritonForge~\cite{li2025tritonforge} utilizes profiling-guided feedback loops to iteratively analyze and identify performance bottlenecks. PRAGMA \cite{lei2025pragma} uses a specialized profiling module to parse low-level quantitative metrics into an interpretable natural language suggestion. KernelBand~\cite{ran2025kernelband} clusters runtime behavior of potential kernels to reduce the exploration space and utilizes profiling data  as context to guide the selection of optimization strategies.

\subsection{Multi-Agent Orchestration}

Recognizing that kernel development inherently involves heterogeneous abilities ranging from algorithmic planning to low-level coding and debugging, recent work increasingly adopts multi-agent designs that explicitly decompose these responsibilities into coordinated roles.

STARK~\cite{dong2025stark} structures generation into Plan-Code-Debug phases to emulate the human team, AKG~\cite{du2025akgkernelagentmultiagent} leverages similar modularity to achieve cross-platform synthesis, while Astra~\cite{wei2025astra} specializes this multi-agent approach for production-grade SGLang kernels. CudaForge~\cite{Zhang2025CudaForge} employs a Coder-Judge framework driven by hardware-level feedback, whereas KForge~\cite{sereda2025kforge} adapts this dual-agent model to new platforms using only single-shot example supervision. Addressing scale, KernelFalcon~\cite{kernelfalcon2024} employs a multi-agent system to tackle the challenge of GPU kernel generation of full machine learning architectures, where the system specifically addresses hierarchical task decomposition and delegation through coordinated manager and worker agents. Moreover, GEAK~\cite{Wang_2025_GEAK} targets AMD GPUs, integrating the generator, reflector, evaluator and optimizer within a Triton-based workflow.

\section{Datasets}
\label{sec:datasets}

\begin{table*}[h!]
\centering
\begin{tabular}{l l p{6.8cm} l}
\toprule
\textbf{Data} & \textbf{Resource} & \textbf{Description} & \textbf{Access} \\ 

\midrule
\multicolumn{4}{l}{\textit{\textbf{I. Structured Datasets (Hugging Face \& Benchmarks)}}} \\ 
\midrule
02/2024 & \textbf{The Stack v2}~\cite{lozhkov2024stackv2} & Unsupervised CUDA/Triton Corpus & \href{https://huggingface.co/datasets/bigcode/the-stack-v2}{[Data]} \\
06/2024 & \textbf{HPC-Instruct}~\cite{hpcinstruct2024} & Instructions for CUDA/MPI/OpenMP & \href{https://huggingface.co/datasets/hpcgroup/hpc-instruct}{[Data]} \\
05/2025 & \textbf{KernelBook}~\cite{kernelbook2025} & Torch-Triton Aligned Corpus & \href{https://huggingface.co/datasets/GPUMODE/KernelBook}{[Data]} \\
02/2025 & \textbf{KernelBench samples} & Kernel Code Snapshots and Profiling Data & \href{https://huggingface.co/datasets/ScalingIntelligence/kernelbench-samples}{[Data]} \\

\midrule
\multicolumn{4}{l}{\textit{\textbf{II. Code-Centric Corpora (GitHub Repositories)}}} \\
\multicolumn{4}{l}{\cellcolor{gray!10}\textit{Layer 1: High-Performance Operator Libraries }} \\ 
12/2017 & \textbf{CUTLASS} & CUDA C++ Template Library for Matrix Ops & \href{https://github.com/NVIDIA/cutlass}{[Code]} \\
05/2022 & \textbf{FlashAttention} & Fast and Memory-Efficient Exact Attention & \href{https://github.com/Dao-AILab/flash-attention}{[Code]} \\
11/2023 & \textbf{FlagAttention} & Memory Efficient Attention Operators in Triton & \href{https://github.com/flagos-ai/FlagAttention}{[Code]} \\
02/2024 & \textbf{AoTriton} & AOT-compiled Triton kernels for AMD ROCm & \href{https://github.com/ROCm/aotriton}{[Code]} \\
11/2021 & \textbf{xFormers} & Hackable and Optimized Transformer Blocks & \href{https://github.com/facebookresearch/xformers}{[Code]} \\
08/2024 & \textbf{Liger-Kernel} & Efficient Triton Kernels for LLM Training & \href{https://github.com/linkedin/Liger-Kernel}{[Code]} \\
04/2024 & \textbf{FlagGems} & Triton-based Operator Library for LLMs & \href{https://github.com/FlagOpen/FlagGems}{[Code]} \\
09/2022 & \textbf{Bitsandbytes} & K-bit Quantization Kernels for LLMs & \href{https://github.com/bitsandbytes-foundation/bitsandbytes}{[Code]} \\
09/2024 & \textbf{Gemlite} & Low-Bit Matrix Multiplication Triton Kernels & \href{https://github.com/dropbox/gemlite}{[Code]} \\
01/2025 & \textbf{FlashInfer} & Kernel Library for Efficient LLM Serving & \href{https://github.com/flashinfer-ai/flashinfer}{[Code]} \\
05/2021 & \textbf{FBGEMM} & Low-Precision Matrix Multiplication & \href{https://github.com/pytorch/FBGEMM}{[Code]} \\
09/2022 & \textbf{Transformer Engine} & Acceleration Library for Transformer Models & \href{https://github.com/NVIDIA/TransformerEngine}{[Code]} \\
09/2025 & \textbf{DeepGEMM} & Clean and Efficient FP8 GEMM Kernels & \href{https://github.com/deepseek-ai/DeepGEMM}{[Code]} \\
04/2026 & \textbf{Tile Kernels} & A Kernel Library Written in TileLang & \href{https://github.com/deepseek-ai/TileKernels}{[Code]} \\

\multicolumn{4}{l}{\cellcolor{gray!10}{\textit{Layer 2: Framework \& System Integration}}} \\
10/2016 & \textbf{PyTorch (ATen)} & Foundational Tensor Library for C++ and Python & \href{https://github.com/pytorch/pytorch}{[Code]} \\
06/2023 & \textbf{vLLM} & High-Efficient Serving Engine & \href{https://github.com/vllm-project/vllm}{[Code]} \\
12/2023 & \textbf{SGLang} & Structured Generation Language for LLMs & \href{https://github.com/sgl-project/sglang}{[Code]} \\
03/2023 & \textbf{llama.cpp} & LLM Inference in C/C++ & \href{https://github.com/ggerganov/llama.cpp}{[Code]} \\
08/2023 & \textbf{TensorRT-LLM} & TensorRT Toolbox for LLM Inference & \href{https://github.com/NVIDIA/TensorRT-LLM}{[Code]} \\
10/2019 & \textbf{DeepSpeed} & System for Large Scale Model Training & \href{https://github.com/deepspeedai/DeepSpeed}{[Code]} \\

\multicolumn{4}{l}{\cellcolor{gray!10}{\textit{Layer 3: Domain-Specific Languages}}} \\
07/2019 & \textbf{Triton} & Open-Source GPU Programming Language & \href{https://github.com/triton-lang/triton}{[Code]} \\
04/2024 & \textbf{TileLang} & Tile-based Optimization Language & \href{https://github.com/tile-ai/tilelang}{[Code]} \\
12/2025 & \textbf{cuTile} & NVIDIA's DSL for Tile-centric Programming & \href{https://docs.nvidia.com/cuda/cutile-python/}{[Link]} \\

\midrule
\multicolumn{4}{l}{\textit{\textbf{III. Knowledge Bases \& Educational Resources}}} \\ 
\multicolumn{4}{l}{\textit{Documentation \& Guides}} \\
06/2007 & \textbf{CUDA Guide} & CUDA C++ Programming Guide & \href{https://docs.nvidia.com/cuda/cuda-c-programming-guide/}{[Docs]} \\
06/2007 & \textbf{PTX ISA} & PTX ISA Reference & \href{https://docs.nvidia.com/cuda/parallel-thread-execution/}{[Docs]} \\
05/2020 & \textbf{Tuning Guides} & NVIDIA Architecture Tuning Guides & \href{https://docs.nvidia.com/cuda/}{[Docs]} \\

\multicolumn{4}{l}{\textit{Community Indices \& Tutorials}} \\
01/2024 & \textbf{GPU-MODE} & Resource Stream \& KernelBook & \href{https://github.com/gpu-mode/resource-stream}{[List]} \\
01/2024 & \textbf{Triton Index} & Community Index for Triton Optimization & \href{https://github.com/gpu-mode/triton-index}{[List]} \\
06/2016 & \textbf{Awesome-CUDA} & Community Curated List for CUDA & \href{https://github.com/Erkaman/Awesome-CUDA}{[List]} \\
12/2023 & \textbf{Awesome-GPU} & Awesome GPU Engineering List & \href{https://github.com/goabiaryan/awesome-gpu-engineering}{[List]} \\
05/2023 & \textbf{LeetCUDA} & CUDA Programming Exercises & \href{https://github.com/xlite-dev/LeetCUDA}{[Code]} \\
01/2023 & \textbf{Triton-Puzzles} & Puzzles for Learning Triton & \href{https://github.com/srush/Triton-Puzzles}{[Code]} \\
01/2011 & \textbf{Colfax Research} & Technical Hub Dedicated to HPC and AI & \href{https://research.colfax-intl.com/}{[Link]} \\
09/2018 & \textbf{Nsight Compute} & Kernel Profiling Guide & \href{https://docs.nvidia.com/nsight-compute/}{[Docs]} \\
07/2024 & \textbf{CUDA Course} & GitHub Repo for CUDA Course & \href{https://github.com/Infatoshi/cuda-course}{[Docs]} \\
\bottomrule
\end{tabular}
\caption{A structured overview of training corpora and kernel knowledge bases.  Note that the dates in the table correspond to the initial release; the libraries themselves continue to undergo active development. }
\label{tab:full_resources}
\end{table*}

The effectiveness of LLM-driven kernel generation depends critically on the availability of domain-specific data. Unlike general software engineering tasks, kernel generation requires models to capture hardware intrinsics, parallel execution semantics, and memory hierarchy constraints. We categorize existing resources into two groups: (1) \textit{Training Corpora}, comprising structured datasets and kernel code repositories used for model training; and (2) \textit{Knowledge Bases}, which typically provide domain-specific knowledge for RAG.

Training corpora include both structured datasets and raw kernel repositories. Structured datasets typically pair high-level intents with optimized implementations, whereas repositories provide collections of expert-written kernels drawn from open-source libraries, training and inference frameworks, and domain-specific languages. Complementing executable code, knowledge bases supply hardware and optimization knowledge through technical documentation, programming guides, tutorials, and community resources. Such knowledge can be incorporated into model pretraining or accessed dynamically through RAG. A comprehensive summary of these resources is provided in Table~\ref{tab:full_resources}. The dates reported correspond to the initial release of each resource.

\section{Benchmark}
\label{sec:benchmarks}
\newcommand{\NVIDIA}{%
  \tikz[baseline=(char.base)]{
    \node[
      circle,
      fill=white!60!white,   
      text=green,            
      inner sep=1.2pt,
      draw=black,            
      font=\sffamily\bfseries\fontsize{6}{7}\selectfont
    ] (char) {N};
  }%
}

\newcommand{\Google}{%
  \tikz[baseline=(char.base)]{
    \node[
      circle,
      fill=white!60!white,   
      text=blue,            
      inner sep=0.6pt,
      draw=black,            
      font=\sffamily\bfseries\fontsize{8}{7}\selectfont
    ] (char) {G};
  }%
}

\newcommand{\HUAWEI}{%
  \tikz[baseline=(char.base)]{
    \node[
      circle,
      fill=white!60!white,   
      text=red,            
      inner sep=0.6pt,
      draw=black,            
      font=\sffamily\bfseries\fontsize{8}{7}\selectfont
    ] (char) {H};
  }%
}

\newcommand{\AMD}{%
  \tikz[baseline=(char.base)]{
    \node[
      circle,
      fill=white!60!white,   
      text=black,            
      inner sep=0.6pt,
      draw=black,            
      font=\sffamily\bfseries\fontsize{8}{7}\selectfont
    ] (char) {A};
  }%
}

\newcommand{\speedup}{%
  \tikz[baseline=(char.base)]{
    \node[
      circle,
      fill=black!60!black,   
      text=white,            
      inner sep=0.6pt,
      draw=blue,            
      font=\sffamily\bfseries\fontsize{8}{7}\selectfont
    ] (char) {S};
  }%
}

\newcommand{\correctness}{%
  \tikz[baseline=(char.base)]{
    \node[
      circle,
      fill=black!60!black,   
      text=white,            
      inner sep=0.6pt,
      draw=blue,            
      font=\sffamily\bfseries\fontsize{8}{7}\selectfont
    ] (char) {C};
  }%
}

\newcommand{\efficiency}{%
  \tikz[baseline=(char.base)]{
    \node[
      circle,
      fill=black!90!black,   
      text=white,            
      inner sep=0.6pt,
      draw=blue,            
      font=\sffamily\bfseries\fontsize{8}{7}\selectfont
    ] (char) {E};
  }%
}

\newcommand{\fastp}{%
  \tikz[baseline=(char.base)]{
    \node[
      circle,
      fill=red!100!red,   
      text=white,            
      inner sep=0.6pt,
      draw=black,            
      font=\sffamily\bfseries\fontsize{8}{7}\selectfont
    ] (char) {f};
  }
}

\newcommand{\perf}{%
  \tikz[baseline=(char.base)]{
    \node[
      circle,
      fill=red!100!red,   
      text=white,            
      inner sep=0.6pt,
      draw=black,            
      font=\sffamily\bfseries\fontsize{8}{7}\selectfont
    ] (char) {P};
  }%
}

\newcommand{\similarity}{%
  \tikz[baseline=(char.base)]{
    \node[
      circle,
      fill=red!100!red,   
      text=white,            
      inner sep=0.6pt,
      draw=black,            
      font=\sffamily\bfseries\fontsize{8}{7}\selectfont
    ] (char) {S};
  }%
}

This section focuses on systematic benchmarking of kernel generation and provides a structured overview of representative evaluation benchmarks, including both evaluation metrics and benchmark datasets, to lay a solid foundation for subsequent method comparison and performance analysis.

\begin{table*}[ht]
\centering
\begin{threeparttable}
\begin{tabular}{p{3.2cm}p{1cm}p{1.2cm}p{1.3cm}p{8.8cm}}
\hline
\textbf{Name} & \textbf{Time} & \textbf{Metrics} & \textbf{Hardware} & \textbf{Description}   \\ \hline
ParEval           &  01/2024  & \correctness \  \speedup \  \efficiency   & \NVIDIA \AMD   &  420 expert-selected tasks across 12 algorithmic domains for benchmarking general parallel code generation.    \\
KernelBench & 02/2025  & \correctness \  \fastp     & \NVIDIA  & 250 PyTorch-to-CUDA kernel generation tasks, curated from popular GitHub repositories and official PyTorch operators, for evaluating AI/DL kernel generation.    \\
TritonBench & 02/2025  & \correctness \speedup \ \similarity  \  \efficiency\tnote{*} & \NVIDIA    & TritonBench evaluates Triton kernel generation via two subsets: 184 high-level kernels sourced from popular GitHub projects (TritonBench-G) and 166 fusion tasks derived from diverse PyTorch operators with different frequencies of usage (TritonBench-T).   \\
MultiKernel-Bench & 07/2025 & \correctness \  \speedup   & \HUAWEI \ \NVIDIA \  \Google   & 285-task benchmark across 14 operator categories for multi-platform DL kernel synthesis.  \\
TritonBench-revised \& ROCm Benchmark & 07/2025  & \correctness \ \speedup   & \AMD   & An AMD GPU-centric evaluation dataset comprising 30 expert-verified ROCm kernels and an adapted version of TritonBench-G, specifically optimized for AMD GPU performance benchmarking.     \\
Robust-kbench  & 09/2025  & \correctness \ \speedup  & \NVIDIA   & A robustness-focused benchmark featuring 9 deep learning task categories, derived by refining and extending KernelBench.  \\
BackendBench & 09/2025 & \correctness \ \speedup & \NVIDIA & A rigorous evaluation framework that enforces PyTorch's official core library standards. Current use cases primarily leverage NVIDIA CUDA and Triton, yet the architecture remains backend-agnostic. \\
CUDAEval      & 10/2025  & \correctness \  \speedup   & \NVIDIA   & Leveraging 313 curated tasks from the Stack v2 to benchmark the efficacy of reasoning transfer in CUDA code optimization.   \\
FlashInfer-Bench & 01/2026  & \correctness \  \fastp     & \NVIDIA  & Provide a unified schema describing kernel definitions, workloads, implementations, and evaluations, including eight representative kernel types used in LLM inference. \\
SOL-ExecBench & 03/2026  & SOL Score     & \NVIDIA  & A benchmark for GPU kernel optimization built around hardware Speed-of-Light (SOL) targets. \\
\hline
\end{tabular}
\begin{tablenotes}
\item[*] Efficiency here is defined as the ratio of the operator's measured throughput to the theoretical maximum performance.
\end{tablenotes}
\end{threeparttable}
\caption{\texorpdfstring{Benchmark datasets for kernel generation and optimization. Metrics: \protect\correctness{} Correctness, \protect\speedup{} Speedup, \protect\efficiency{} Efficiency, \protect\fastp{} fast$_p$, \protect\similarity{} Similarity. Hardware Platforms: \protect\NVIDIA{} NVIDIA GPUs, \protect\HUAWEI{} HUAWEI NPUs, 
\protect\Google{} Google TPUs, \protect\AMD{} AMD GPUs.}{Benchmark datasets for code generation and optimization in kernel generation. 
Metrics: Correctness, Speedup, Efficiency, fast$_p$, Perf, Similarity. Hardware Platforms: NVIDIA GPUs, HUAWEI NPUs, Google TPUs, AMD GPUs.}}
\label{tab:benchmarks}
\end{table*}

\subsection{Metrics}

Kernel evaluation typically considers three aspects: correctness, performance, and composite quality metrics. Most benchmarks adopt execution-based testing, where generated kernels are executed and compared against reference CUDA or PyTorch implementations. Since kernel generation is inherently stochastic, evaluations are often repeated across multiple samples and random seeds.

\textbf{Correctness} measures whether a generated kernel produces outputs consistent with a reference implementation within predefined numerical tolerances. Evaluation criteria vary across kernel types and precision formats (e.g., FP16, BF16, and FP8), reflecting their distinct numerical characteristics. Detailed evaluation protocols are provided in FlashinferBench~\cite{xing2026flashinfer}.

\textbf{Performance} is primarily measured by runtime speedup relative to a reference implementation, typically using wall-clock execution time averaged over repeated trials. Some benchmarks further compare achieved performance against Speed-of-Light (SOL) estimates, which approximate the theoretical efficiency limit of the target hardware.

In addition, Efficiency refers to how effectively the generated operators utilize computation resources during execution, and Compatibility is considered when evaluating operator generation techniques across different hardware platforms or languages. Composite metrics jointly assess multiple dimensions of kernel quality. Representative examples include \textit{Similarity}, which evaluates code similarity using lexical, syntactic, and dataflow features; and \textit{fast$_p$}, which combines correctness and performance by measuring the proportion of generated kernels that are both correct and achieve a speedup greater than $p$:
\begin{equation}
\text{fast$_p$} = \frac{1}{N}\sum_{i=1}^N \textbf{1}(correct_i \land \{speedup_i > p\} ) .
\end{equation}

\subsection{Benchmark Datasets}

As summarized in Table~\ref{tab:benchmarks}, kernel benchmarks are progressively evolving toward more realistic and comprehensive evaluation. This evolution is reflected in three dimensions: \emph{metrics}, where evaluation has expanded from correctness and runtime speedup (ParEval~\cite{pareval2024}) to comprehensive measures of efficiency, robustness, and overall kernel quality. Examples include fast$_p$ in KernelBench, efficiency metrics in TritonBench~\cite{Li_2025_TritonBench}; \emph{hardware coverage}, where benchmarks increasingly extend beyond NVIDIA GPUs to encompass AMD GPUs, Huawei NPUs, and Google TPUs (such as MultiKernelBench~\cite{wen2025multikernelbench}); and \emph{workload diversity}, where evaluation has shifted from simple operators toward production-grade kernels derived from real-world frameworks, systems, and AI workloads. For example, FlashInfer-Bench and SOL-ExecBench~\cite{lin2026sol} focus on critical kernels derived from production systems and emerging AI models, and BackendBench~\cite{saroufim2025backendbench} targets complex edge cases.

\section{Challenges and Opportunities}
\label{sec:future_work}

Despite rapid progress, LLM-driven kernel generation remains a nascent research area. Advancing from proof-of-concept systems to robust and scalable deployment requires addressing challenges spanning evaluation, data, agent capabilities, infrastructure, and human-AI collaboration. This section reviews these challenges and outlines promising directions for future research.

\paragraph{Evaluation Reliability and Generalization.} 
Reliable evaluation remains a fundamental challenge for LLM-driven kernel generation. Existing systems are vulnerable to reward hacking, where kernels achieve favorable benchmark scores without delivering corresponding benefits in practical deployments. At the same time, current benchmarks often cover limited workloads, hardware platforms, and execution settings, raising concerns about the generalization of reported methods. Furthermore, the impact of kernel-level improvements on end-to-end AI systems remains insufficiently understood. Future evaluation frameworks should emphasize robustness against reward hacking, broad generalization across workloads or platforms, and system-level assessment, providing a more reliable measure of real-world effectiveness.

\paragraph{Data Scarcity and Synthetic Scaling.} 
Data scarcity remains a major bottleneck for LLM-driven kernel generation. High-performance kernels are sparsely represented in existing corpora, and predominantly contain final implementations while largely omitting optimization trajectories and hardware-aware expertise. As a result, critical signals for learning kernel optimization remain limited. Promising directions include large-scale kernel dataset construction, synthetic data generation, and the collection of execution-driven optimization traces. Such resources could support both model training and agentic learning, and may be essential for scaling kernel generation capabilities.

\paragraph{Agentic Training and Harness Engineering.} 
Kernel optimization poses a challenging long-horizon task that requires reasoning over iterative cycles of generation, execution, profiling, and refinement. Current foundation models are not explicitly trained for such long-horizon optimization trajectories, while existing agentic systems often rely on handcrafted workflows that struggle with exploration efficiency, context management, and long-term credit assignment. Future progress may require advances in both long-horizon agentic training and autonomous harness engineering.
Moreover, recent advances in general-purpose coding agents (e.g., Claude Code, OpenCode) suggest that kernel optimization may not require developing task-specific agents from scratch. Instead, foundation agents can be specialized through harness engineering, leveraging domain-specific environments, tools, and feedback to enable scalable agentic performance engineering.

\paragraph{Scalable Infrastructure for Synthesis and Training.}
Scalable infrastructure remains a critical bottleneck for large-scale data synthesis and agentic training. Key challenges include building distributed and securely isolated sandbox environments for execution, addressing latency mismatches between agent rollout and kernel compilation and verification, and designing fault-tolerant, multi-device services that ensure stable and efficient training at scale.
Advances in such infrastructure are essential for transforming kernel synthesis and data sampling from low-throughput, ad-hoc experimentation into a systematic, data-driven evolutionary process, enabling continuous improvement of agentic kernel optimization systems.

\paragraph{Human-AI Collaboration for Kernel Generation.}

Human-AI collaboration represents a promising alternative to fully autonomous kernel optimization. A central challenge is to establish effective feedback loops between human expertise and agentic exploration. While human-in-the-loop guidance can guide optimization through high-level objectives and constraints, human-from-the-loop learning enables knowledge discovered by agents to be transferred back to developers. Given the verifiable nature of kernel optimization, such bidirectional collaboration may create an effective cycle of human and agent co-evolving, expanding the frontier of automated performance engineering.

\section{Conclusion}
\label{sec:conclusion}

This survey highlights the transformative potential of LLMs and agentic systems for automating kernel generation and optimization, synthesizing recent advances in methodologies alongside the development of kernel-centric datasets and benchmarks. Looking ahead, future progress will depend on more reliable evaluation protocols, as well as the data infrastructure and harness engineering required for agentic kernel generation. These advances have the potential not only to alleviate the burden of manual kernel engineering, but also to unlock substantial productivity gains for rapidly scaling AI infrastructure.





\bibliographystyle{named}
\bibliography{ijcai26}

\end{document}